\documentclass[final,5p,times,twocolumn]{elsarticle}

\usepackage{amssymb}
\usepackage{amsmath}
\usepackage{siunitx}
\usepackage{tabularx}
\usepackage{algorithmic}

\usepackage{algorithm2e} 
\usepackage{amsthm}
\usepackage{graphicx}
\usepackage{lscape}
\usepackage{verbatim}
\usepackage{color,soul}
\usepackage{xcolor}
\usepackage{subfigure}
\usepackage{tabularx,ragged2e,booktabs,caption,array,multirow,multicol}
\usepackage{csquotes}
\usepackage{mathtools}
\usepackage{lineno}
\usepackage{amsthm}  
\usepackage{listings}
\usepackage{amssymb}
\usepackage{latexsym}
\usepackage{epsfig}
\usepackage{float}
\usepackage{xspace} 

\usepackage{subfigure} 

\usepackage{tikz}
\usetikzlibrary{shapes,arrows}
\usepackage{xcolor} 
\usepackage{tikz}
\usetikzlibrary{fit,calc}
\newcommand*{\tikzmk}[1]{\tikz[remember picture,overlay,] \node (#1) {};\ignorespaces}
\newcommand{\boxit}[1]{\tikz[remember picture,overlay]{\node[yshift=3pt,fill=#1,opacity=.25,fit={(A)($(B)+(.95\linewidth,.8\baselineskip)$)}] {};}\ignorespaces}
\colorlet{pink}{red!40}
\colorlet{blue}{cyan!60}

\journal{Eng. App. AI}

\begin{document}

\begin{frontmatter}

\title{ Surrogate-assisted parallel tempering for Bayesian neural learning }

\author[geo]{Rohitash Chandra}
\ead{rohitash.chandra@sydney.edu.au}
\ead[url]{rohitash-chandra.github.io}

\author[iit,ctds]{Konark Jain}
\ead{konark145@gmail.com }

\author[ctds,srm]{Arpit Kapoor}
\ead{kapoor.arpit97@gmail.com}

\author[ctds,iitdel]{Ashray Aman}
\ead{kapoor.arpit97@gmail.com}

\address[geo]{School of Mathematics and Statistics, University of New South Wales, Sydney, 
NSW 2052, Australia}

\address[ctds]{Centre for Translational Data Science, The University of Sydney, 
NSW 2006, Australia}

\address[srm]{Department of Computer Science and Engineering, SRM Institute of Science and Technology, Chennai, Tamil Nadu, India}

\address[iit]{ Department of Electronics and Electrical Engineering, Indian Institute of Technology Guwahati, Assam, India }

\address[iitdel]{ Department of Mathematics, Indian Institute of Technology Delhi, Delhi, India }


\begin{abstract}
Due to the need for robust uncertainty quantification, Bayesian neural learning has gained attention in the era of deep learning and big data. Markov Chain Monte-Carlo (MCMC) methods typically implement Bayesian inference which faces several challenges given a large number of parameters, complex and multimodal posterior distributions, and computational complexity of large neural network models.  Parallel tempering MCMC addresses some of these limitations given that they can sample multimodal posterior distributions and utilize high-performance computing. However, certain challenges remain given large neural network models and big data.    Surrogate-assisted optimization features  the estimation of an objective function  for models which are computationally expensive. In this paper, we address the inefficiency of parallel tempering MCMC for large-scale problems by combining parallel computing features   with surrogate assisted likelihood estimation  that describes the plausibility of a model parameter value, given specific observed data.  Hence, we present surrogate-assisted parallel tempering for Bayesian neural learning for simple to computationally expensive models. Our results demonstrate that the methodology significantly lowers the computational cost while maintaining quality in decision making with Bayesian neural networks. The method has applications for a Bayesian inversion and uncertainty quantification for a broad range of numerical models.

\end{abstract}



\end{frontmatter}

\section{Introduction}

 Although neural networks have gained significant attention due to the deep learning revolution \cite{schmidhuber2015deep}, several limitations exist. The challenge widens for uncertainty quantification in decision making given the development of new neural network architectures and learning algorithms.   Bayesian neural learning provides a probabilistic viewpoint with the representation of neural network weights as probability distributions \cite{mackay1995probable,robert2014machine}.   
Rather than single-point estimates by gradient-based learning methods, the probability distributions 
 naturally account for uncertainty in parameter estimates. 
 
Through Bayesian neural learning, uncertainty regarding the data and model can be propagated into the decision making process.  Markov Chain Monte Carlo sampling   methods (MCMC) implement Bayesian inference \cite{hastings1970monte,metropolis1953equation,tarantola1982inverse,mosegaard1995monte} by constructing a  Markov chain, such that the desired distribution becomes the equilibrium distribution after a number of steps \cite{raftery1996,van2016simple}. MCMC  methods provide numerical approximations of multi-dimensional integrals  \cite{banerjee2014hierarchical}.    MCMC methods have not gained as much attention in neural learning when compared to gradient-based counterparts since convergence becomes computationally expensive for a large number of model parameters and multimodal posteriors \cite{robert2018accelerating}.  MCMC methods typically require thousands of samples to be drawn depending on the model which becomes a major limitation in applications such as deep learning  \cite{schmidhuber2015deep,gal2016dropout,kendall2017uncertainties}. \textcolor{black}{Hence, other methods for implementing Bayesian inference exist, such as variational inference \cite{blei2017variational,damianou2016variational} which has been used for deep learning \cite{blundell2015weight}.}   Variational inference has the advantage of faster convergence when compared to MCMC methods for large models. However, in computationally expensive models, variational inference methods would have a similar problem as MCMC,  since both would need to evaluate model samples for the likelihood;  hence, both would benefit from surrogate-based models.

 Parallel tempering is   an MCMC method that  \cite{swendsen1987nonuniversal,marinari1992simulated,geyer1995annealing} features multiple replicas to provide global and local exploration   which makes them suitable for irregular and multi-modal distributions \cite{patriksson2008temperature,hukushima1996exchange}. \textcolor{black}{During sampling, parallel tempering features the exchange of neighbouring replicas to feature   exploration and exploitation in the search space.}  The replicas with higher temperature values make sure that there is enough exploration, while replicas with lower temperature values  exploit the promising areas found during the exploration.  In contrast to canonical MCMC sampling methods, parallel tempering is more easily implemented in a multi-core or parallel computing architecture \cite{lamport1986interprocess}.  In the case of neural networks, parallel tempering was used for inference of restricted 
 Boltzmann machines (RBMs) \cite{salakhutdinov2007restricted,FISCHER2015102} where it was shown that  \cite{desjardins2010tempered} parallel tempering is more effective than Gibbs sampling  by itself.  Parallel tempering for RBMs has been improved by featuring the efficient exchange of information among the replicas \cite{brakel2012training}. These studies motivated   the use of parallel tempering in Bayesian neural learning for pattern classification and regression tasks  \cite{Chandra2019NC}.

 Surrogate assisted optimisation \cite{hicks1978wing,jin2011surrogate} considers the use of machine learning methods such as Gaussian process and neural network models to estimate the objective function during optimisation. This is handy when the evaluation of the objective function is too time-consuming.  In the past, metaheuristic and  evolutionary optimization methods have been used in surrogate assisted optimization \cite{ong2003evolutionary,zhou2007combining}. Surrogate assisted optimization has been useful for the fields of  engine  and aerospace design  to   replicate  computationally expensive models \cite{ong2005surrogate,jeong2005efficient,samad2008multiple,hicks1978wing}. The optimisation literature motivates improving parallel tempering using a low-cost replica of the actual model via a surrogate to lower the computational costs. 
  
 In the case of conventional Bayesian neural learning, much of the literature concentrated on smaller problems such as datasets and network architecture \cite{richard1991neural,mackay1996hyperparameters,mackay1995probable,robert2014machine}  due to  computational efficiency of MCMC methods. Therefore, parallel computing has been used in the implementation of parallel tempering for Bayesian neural learning \cite{Chandra2019NC}, where the computational time was significantly decreased due to parallelisation. Besides, the method achieved better prediction accuracy and convergence due to the exploration features of parallel tempering. We believe that this can be further improved through incorporating notions from surrogate assisted optimisation in parallel tempering, \textcolor{black}{where the likelihood function at certain times is estimated rather than evaluated in a high-performance computing environment}.

 We note that some work has been done using surrogate assisted Bayesian inference.    Zeng \emph{et. al} \cite{wang2016surrogate}  presented a method for material identification using surrogate assisted Bayesian inference for estimating the parameters of advanced high strength steel used in vehicles. Ray  and Myer  \cite{Ray2019} used Gaussian process-based surrogate models with MCMC for geophysical inversion problems. The benefits of surrogate assisted methods for computationally expensive optimisation problems motivate parallel tempering for computationally expensive models. 
  To our knowledge, there is no work on parallel tempering with surrogate-models implemented via parallel computing for machine learning problems.    In the case of parallel tempering that uses parallel computing, the challenge would be in developing a paradigm where different replicas can communicate efficiently. Besides,  the task of training the surrogate model from data across multiple replicas in parallel poses further challenges.

 In this paper,  we present surrogate-assisted parallel tempering for Bayesian neural learning where a surrogate is used to estimate the likelihood rather than evaluating the actual model that feature a large number of parameters and datasets. We present a framework that seamlessly incorporates the decision making by a master surrogate for parallel processing cores that execute the respective replicas of parallel tempering MCMC. Although the framework is intended for general computationally expensive models,  we demonstrate its effectiveness using a neural network model for classification problems.  The major contribution of this paper is to address the limitations of parallel tempering given computationally expensive models.

 The rest of the paper is organised as follows. Section 2 provides background and related work, while Section 3 presents the proposed  methodology. Section 4 presents experiments and results and Section 5 concludes the paper with discussion for future research. 
 
 \section{Related work}

 \subsection{Bayesian neural learning}
  
 In  Bayesian inference, we update the probability for a hypothesis as more evidence or information becomes available \cite{freedman1963asymptotic}.  We estimate the posterior distribution by sampling using prior distribution and a 'likelihood function' that evaluates the model with observed data.  A probabilistic perspective treats learning as equivalent to maximum likelihood estimation (MLE)  \cite{white1982maximum}.  Given that the neural network is the model, we base the prior distribution on belief or expert opinion without observing the evidence or training data \cite{richard1991neural}. An example of information or belief for the prior in the case of neural networks  is the concept of \textit{weight decay} that states  that smaller weights are better for generalization \cite{krogh1992simple,mackay1995probable,mackay1995probable,neal2012bayesian,auld2007bayesian}. 
 
   Due to limitations in  MCMC sampling methods,   progress in development and applications of Bayesian neural learning has been slow, especially when considering larger neural network architectures, big data and deep learning. Several techniques have been applied to improve MCMC sampling methods by incorporating approaches from the optimisation literature.     Neal \emph{et al.} \cite{neal2011mcmc} presented Hamiltonian dynamics  that involve using   gradient information for constructing efficient MCMC proposals during sampling. Gradient-based learning using Langevin dynamics refer to use of gradient information with Gaussian noise \cite{welling2011bayesian}; Chandra \emph{et al.} employed Langevin dynamics for  Bayesian neural networks for time series prediction \cite{Chandra2017Langevin}. Hinton \emph{et al.}\cite{hinton2006fast} used complementary priors  for deep belief networks  to  form an undirected associative memory for  handwriting recognition. Furthermore, parallel tempering has been used for the  Gaussian Bernoulli Restricted Boltzmann Machines (RBMs) \cite{cho2011improved}. Prior to this, Cho \emph{et al.} \cite{cho2010parallel} demonstrated the efficiency of parallel tempering in RBMs. Desjardins \emph{et al.} \cite{desjardins2010adaptive} utilized parallel tempering for maximum likelihood training of RBMs   and later used it for deep learning using RBMs \cite{desjardins2014deep}.

In Bayesian neural learning, we estimate the posterior distribution by MCMC sampling using a 'likelihood function' that evaluates the model given the observed data and prior distribution. Given input features or covariates ($\bf x_t$),  we compute  $f({\bf x}_t)$ by a feedforward neural network with one hidden layer,

\begin{equation}
f({\bf x}_t)   =   g \bigg(  \delta_o + 
\sum_{h=1}^{H} v_{ho} g \bigg(  \delta_h + \sum_{d=1}^{I} (w_{dh} 
\bf x_{t} )\bigg) 
\label{expected_y}
\end{equation}

\noindent where $\delta_o$ and $\delta_h$  are the bias  for the output $o$ and 
hidden $h$ layer, respectively.  $v_{ho}$ is the weight which maps the hidden 
layer $h$ to the output layer. $w_{dh}$ is the weight which maps $\bf x_{t}$ to 
the hidden layer $h$ and $g$ is the activation function 
 for the hidden and output layer units. 
 
 Let $\boldsymbol{\theta}= 
( \mathbf {w},\mathbf {v}, \boldsymbol  {\delta}, \tau^2)$, with $ \boldsymbol  {\delta}=(\delta_o,\delta_h)$,
and  $\mathcal{L}$ as the number of  parameters that includes weights and 
biases. $\tau^2$ is a single parameter to represent the noise in the predictions given by the neural network model.   $I, H, O$ refers to number of input, hidden and output neurons, respectively.   We assume a Gaussian  prior distribution  using,

\begin{eqnarray}
\log\left(p(\boldsymbol{\theta})\right)&=&-\frac{\mathcal{L}}{2}\log(\hat{\sigma}^2)\nonumber\\
&&-\frac{1}{2\hat{\sigma}^2}\left(\sum_{h=1}^H\sum_{d=1}^I w_{dh}^2+\sum_{h=1}^H(\delta_h^2+v_{h}^2)+\delta_o^2\right)  \nonumber\\
\label{eq:prior}
\end{eqnarray}

\noindent where the variance ($\sigma^2$) is user-defined, gathered by information (prior belief) regarding the distribution of weights of trained neural networks in similar applications.

\textcolor{black}{Given data ($\bf y$), we use Bayes rule for  the posterior $p(\boldsymbol{\theta}|{\bf y})$  which is proportional to the likelihood  $p({\bf y}|\boldsymbol{\theta})$  times the prior $p(\boldsymbol{\theta}) $. }
\[
  p(\boldsymbol{\theta}|{\bf y})  \propto  p({\bf y}|\boldsymbol{\theta})  \times p(\boldsymbol{\theta})  
\]

\noindent Then, the log-posterior transforms to

\[
\log\left(p(\boldsymbol{\theta}|{\bf y})\right)=\log\left(p(\boldsymbol{\theta})\right)+\log\left(p({\bf y}|\boldsymbol{\theta})\right)
\]
 
\noindent  To implement the likelihood given in Equations 1 and 2, we use the multinomial likelihood function for the classification problems as shown in Equation \ref{multinomial}.

\begin{eqnarray}
\log\left(p({\bf y }|\boldsymbol{\theta})\right)&=&\sum_{t\in T}\sum_{k=1}^K z_{t,k}\log{\pi_k}
\label{multinomial}
\end{eqnarray}
\noindent for classes $ k = 1,\ldots,K$, where $\pi_k$ is the output  of the neural network  after applying the transfer function, and $T$ is the number of training samples.  In this case, the transfer function    is the softmax function \cite{bishop1995neural},
\begin{equation}
\pi_k = \frac{\exp(f(x_p))}{\sum_{k=1}^K \exp(f(x_k))}
\end{equation}
\noindent for $k = 1,\ldots,K$. $z_{t,k}$ is an indicator variable for the given sample $t$ and the class $k$ as given in the dataset and  defined by,  
\begin{equation}
z_{t,k} =
\begin{cases}
    1,& \text{if } y_t = k\\
    0,              & \text{otherwise}
\end{cases}
\end{equation}

\subsection{Parallel tempering MCMC}

 Parallel tempering MCMC features an ensemble of chains (known as replicas)   executed at different {\it temperature levels} defined by a {\it temperature ladder} that determine the extent of  exploration and exploitation  \cite{swendsen1986replica,hukushima1996exchange,hansmann1997parallel,sambridge2014parallel}.  The replicas with higher temperature values ensure that there is enough exploration, while replicas with lower temperature values exploit the promising areas found by exploration. Typically, the neighbouring replicas are exchanged given the Metropolis-Hastings criterion. In some implementations, non-neighboring replicas are  also considered for exchange  \cite{sambridge2014parallel,ray2016frequency}. In such cases, we calculate the acceptance probabilities of all possible moves a \textit{priori}. The specific exchange move is then selected, which is useful when a limited number of replicas are available \cite{calvo2005all}.  Although geometrically uniform temperature levels have been typically used for the respective replicas, determining the optimal tempering assigned for each of the replicas has been a challenge that attracted some attention in the literature \cite{rathore2005optimal,katzgraber2006feedback,bittner2008make,patriksson2008temperature}. Typically, gradient-free proposals within chains are used for proposals for exploring multimodal and discontinuous posteriors \cite{sen1996bayesian,maraschini2010monte}; however, it is possible to incorporate gradient-based information for developing effective proposal distributions \cite{Chandra2019NC}. 
 
 Although denoted "parallel", the replicas can be executed sequentially in a single processing unit  \cite{ray2013robust}; however, multi-core or high-performance computing systems can feature parallel implementations improving the computational time \cite{ray2016frequency,ray2018low, LI2009269}. There are challenges since parallel tempering features exchange or transition between neighbouring replicas, and we need to consider efficient strategies that take into account inter-process communication  \cite{LI2009269}.  Parallel tempering has been used with high-performance computing for seismic full waveform inversion (FWI) problems \cite{ray2018low}, which is amongst the most computationally intensive problems in earth science today.
 
  Parallel tempering has also been implemented in a distributed volunteer computing network via crowd-sourcing for multi-threading and graphic processing units \cite{karimi2011high}. Furthermore, implementation with field-programmable gate array (FPGA) has shown much better performance than multi-core and graphic processing units (GPU) implementations \cite{mingas2017particle}.

  In a parallel tempering MCMC ensemble, given $M$ replicas defined by multiple temperature levels, the state of the ensemble is specified by   $X= {x_1,x_2, ..., x_M}$; where  $x_i$   is the $i$th replica, with   temperature level $T_i$. The equilibrium distribution of the ensemble $X$ is given by,
 
\begin{equation}
\Pi(X) = \prod_{i=1}^{M} \frac{ \exp( - \frac{1}{T_i}   L(x_i))}  {Z(T_i)  }
\end{equation}

\noindent where {$L(x_i)$ is the   log-likelihood function}   for the replica state at each temperature level ($T_i$) and $Z(T_i)$ is an intractable normalization constant. At every iteration of the replica state, the replica can feature two types of transitions that includes a  \textit{Metropolis transition} and a \textit{replica transition}. 

The \textit{Metropolis transition} enables the replica to perform local Monte Carlo moves  defined by the   energy function $E(x_i)$. The  local replica state  $x^*_i$ is sampled using  a proposal distribution $q_i(.|x_i)$ and the Metropolis-Hastings acceptance probability $\alpha$ for the local replica  $L_{local}$   is given as,

\begin{equation}
\alpha = min \Bigg(1, \exp \Big(- \frac{1}{T_i}   (L(x^*_i) - L(x_i))\Big)  \Bigg)
\end{equation}

\noindent  The detailed balance condition holds for each replica, and therefore it holds for the ensemble system.

Typically, the Metropolis-Hastings update consists of a single stochastic process that evaluates the energy of the system and accepted is based on the temperature ladder.  The selection of the temperature ladder for the replicas is done prior to sampling;  we use a geometric spacing methodology  \cite{vousden2015dynamic} as given,

\begin{equation} 
 T_i = T_{\max}^{\frac{(i-1)}{(M-1)}}
 \label{eq:geometric}
 \end{equation}
 
  \noindent where  $i = 1, \ldots, M$ and $T_{\max}$ is the maximum temperature which is user-defined.

The \textit{Replica transition}  features  the exchange of two neighbouring replica states  ( $x_i \leftrightarrow x_{i+1}$) defined by their  temperature level, $i$ and $i +1$. The replica exchange is accepted by the Metropolis-Hastings criterion with replica exchange  probability $\beta$ by,

\begin{equation}  
\beta = min \bigg(1, \exp \bigg( \bigg(\frac{1}{T_{i+1}}   - \frac{1}{T_i}    \bigg)  \bigg(L(x_{i+1}) -  L(x_{i})\bigg) \bigg)
 \label{eq:swap}
 \end{equation}

 \noindent Based on the Metropolis criterion, the configuration  (position in the replica )  of neighbouring replicas at different temperatures are exchanged. This results in a   robust ensemble which can sample both low and high energy configurations. 

The replica-exchange enables a replica that could be stuck at a local minimum with low-temperature level to exchange configuration with a higher neighbouring temperature level and hence improve exploration. In this way, the replica-exchange can shorten the sampling time required for convergence. The frequency of determining the exchange and the temperature level is user-defined. For further details about the derivation for the equations in this section, see \cite{sambridge2014parallel,earl2005parallel}.

 \subsection{Surrogate-assisted optimization}

 Surrogate assistant optimisation refers to the use of machine learning or statistical learning models to develop approximately computationally inexpensive simulation of the actual model \cite{jin2011surrogate}. The major advantage is that the surrogate model  provides computationally efficiency when compared to the exact model used for evolutionary algorithms and related optimisation methods \cite{ong2003evolutionary,zhou2007combining}.  In the optimisation literature, such approaches are also known as response surface methodologies \cite{Douglas1977,letsinger1996response} which have been applicable for a wide range of engineering problems such as reliability analysis of laterally loaded piles  \cite{tandjiria2000reliability}.

 In the case of evolutionary computation methods, Ong \emph{et al.} \cite{ong2003evolutionary}   presented parallel evolutionary optimisation for solving computationally expensive functions with application to aerodynamic wing design where surrogate models used radial basis functions. Zhou \emph{et al.} \cite{zhou2007combining}  accelerated evolutionary optimization  with   global and local surrogate models while Lim \emph{et al.}
 \cite{lim2010generalizing} presented a generalised method that accounted for uncertainty in estimation to unify diverse surrogate models. Jin \cite{jin2011surrogate}  reviewed surrogate-assisted evolutionary computation that covered single and multi-objective,  dynamic, constrained,  and multimodal optimisation problems. Furthermore, surrogate models have been widely used in Earth sciences such as modelling water resources \cite{razavi2012review}. Moreover, Díaz-Manríquez \emph{et al.}  \cite{diaz2016review}  presented a review of surrogate assisted multi-objective evolutionary algorithms that showed that the method has been successful in a wide range of application problems. 
 

 The search for the right surrogate model is a significant challenge given different types of likelihood or fitness landscape given by the actual model. Giunta \emph{et al.} \cite{giunta1998comparison} presented a comparison of quadratic polynomial models with the least-square method and interpolation models that featured   Gaussian process regression (kriging). They discovered that the quadratic polynomial models were more accurate in terms of errors for estimation for the optimisation problems. Jin \emph{et al.} \cite{jin2001comparative} presented another study that compared several surrogate models that include polynomial regression, multivariate adaptive regression splines, radial-basis functions, and kriging based on multiple performance criteria using different classes of problems. The authors reported radial basis functions as one of the best for scalability and robustness, given different types of problems. They also reported kriging to be computationally expensive. 
   
  \section{Surrogate-assisted  multi-core parallel  tempering}

Surrogate models primarily learn to mimic actual or true models using their behaviour, i.e. how the  true model responds to a set of input parameters. A surrogate model captures the relationship between the input and output given by the true model. The input is the set of proposals in parallel tempering  MCMC that features the weights and biases of the neural network model.  Hence,  we utilize the surrogate model to approximate the likelihood of the true model. We define the approximation of the likelihood by the surrogate as \textit{pseudo-likelihood}. We train the surrogate model on the data that is composed of the history of proposals for weights and biases of the neural network with the corresponding true likelihood. We do not estimate the output of the neural network by the surrogate (since in some problems there are many outputs); hence to limit the number of variables, we directly estimate the likelihood using the surrogate.

We implement the neighbouring replica transition or exchange at regular intervals.  The cost of inter-process communication must be limited to avoid computational overhead, given that we execute each replica on a separate processing core. The  \textit{swap interval} defines the time (number of iterations or samples) after which each replica pauses and awaits for neighbouring replica exchange. After the exchange, the replica manager process enables local  Metropolis transition and the process repeats. We note that although we use a neural network model, the framework is general and we can use other models; which could include those from other domains such as  Earth science models that are computationally expensive \cite{chandra2019PT-Bayeslands,sambridge2013parallel}.

 Bayesian neural learning employs parallel tempering MCMC for inference; this can be viewed as a training procedure for the neural network model. The goal of the surrogate model is to save computational time taken for evaluation of the true likelihood function associated with computationally expensive models.

Given that the  true model is represented as $L = f(x)$, the surrogate model provides an approximation in the form $\hat{l} = \hat{f}(x) $, such that $L = \hat{l} + e$; where  $e$ represents the difference or error. The surrogate model   provides an estimate by the  \textit{pseudo-likelihood}  for replacing  \textit{true-likelihood} when needed. The surrogate model is constructed by training from experience which is given by the set of input $\bf{x}_{i,s}$ with corresponding true-likelihoods $L_{i,s}$; where $s$ represents the sample and $i$ represents the replica.  Hence,  input features ($\Phi$) for the surrogate is developed by combining  $\bf{x}_{i,s}$  using samples generated ($\theta$) across all the replicas for a given surrogate interval ($\psi$).   The surrogate interval defines the batch size of the training data for the surrogate model which goes through incremental learning. All the respective replicas sample until the surrogate interval is reached and then the manager process collects the sampled data to create a training data batch for the surrogate model. This can be formulated as follows,

\begin{eqnarray}
\Phi&=&([\bf{x}_{1,s},\ldots,\bf{x}_{1,s+\psi}], \ldots, [ \bf{x}_{M,s},\ldots,\bf{x}_{M,s+\psi}])\nonumber\\
\lambda &=&([L_{1,s},\ldots,L_{1,s+\psi}], \ldots,  [L_{M,s},\ldots,L_{M,s+\psi}])\nonumber \\
\Theta &=& [\Phi, \lambda]
\label{data}
\end{eqnarray}

where $\bf{x}_{i,s}  $  represents the set of parameters proposed  and  $s$, $y_{i,s}$ is the output from the multinomial  likelihood, and $M$ is the total number of replicas. 

The training surrogate dataset ($\Theta = [\Phi, \lambda]$) consists of input features ($\Phi$) and response ($ \lambda$) for  the span of each surrogate interval  ($s+\psi$). Hence, we  denote the pseudo-likelihood $\hat{y}$  by $\hat{y} = \hat{f}(\Theta) $; where $\hat{f}$ is the surrogate model.
 We amend the likelihood in training data for the temperature level since it has been changed by taking ${L_{local}}/{T_i}$ for given replica $i$.  The likelihood is amended to reflect the true likelihood rather than that represented by taking into account the temperature level. All the respective replica $\Theta_i$  data is  combined, $\Upsilon = [\Theta_1, \Theta_2, ... \Theta_N]$ and trained using the neural network model given in Equation  \ref{expected_y}.

Algorithm \ref{alg:ptfnn} presents    surrogate-assisted multi-core parallel tempering   for Bayesian neural learning. We implement the algorithm using (multi-core) parallel processing where a manager process takes care of the ensemble of replicas that run in separate processing cores. Given the parallel processing nature, it is tricky to implement when to terminate sampling distributed amongst parallel processing replicas; hence, our termination condition waits for all the replica processes to end. We monitor the number of \textit{alive replica process} in the master process by setting the number of alive replicas in the ensemble   ($alive = M$). We note that the highlighted region of Algorithm 1 shows different processing cores as given in Figure \ref{fig:surrogate}. We highlight the manager process in blue, and in pink we highlight the ensemble of replica processes running in parallel. We initially assign the replicas that sample $\theta_n$  with values using the Gaussian prior distribution with user-defined variance ($\sigma^2=25$) centred at the mean of 0.    We define the temperature level  by geometric ladder (Equation \ref{eq:geometric}), and  other key parameters  which includes the number of replica samples   ($R_{max}$), 
swap-interval which defines after how many samples to check for replica swap  ($R_{swap}$),  surrogate interval  ($\psi$), and surrogate probability  ($S_{prob}$).  The main purpose of the surrogate interval is to collect enough data for the surrogate model during sampling. This also can be seen as batch-based training where we update the model after we collect  the data at regular intervals.

  Figure \ref{fig:surrogate} shows how we use the manager processing unit for the respective replicas running in parallel for the given surrogate interval.  The manager process waits for all the replicas to reach the surrogate interval. Then we calculate the replica transition probability for the possibility of swapping the neighbouring replicas. Figure \ref{fig:surrogate}  further highlights the information flow between the master process and the replica process via inter-process communication    \footnote{Python multiprocessing library for implementation: https://docs.python.org/2/library/multiprocessing.html}.  The information flows from the replica process to master process using \textit{signal()} given by the replica process as shown in Stage 2.2 and 5.0 of Algorithm 1.

 The surrogate model is re-trained for remaining surrogate interval blocks until the maximum iteration ($R_{max}$) is reached to enable better estimation for the pseudo-likelihood.  The surrogate model is trained only in the manger process, and we pass a copy of the surrogate model with the trained parameters to the ensemble of replica processes for estimating the pseudo-likelihood when needed.  Note that only the samples associated with the true-likelihood become part of the surrogate training dataset. The surrogate training can consume a significant portion of time which is dependent on the size of the problem in terms of the number of parameters and the surrogate model used. We evaluate the trade-off between quality of estimation by pseudo-likelihood and overall cost of computation for the true likelihood function for different types of problems. 
 
 In Algorithm 1, Stage 1.4 predicts the pseudo-likelihood ($L_{surrogate}$) with given proposal $\theta_s^*$. Stage 1.5 calculates the likelihood moving average of past three likelihood values,  $L_{past}$ = mean($L_{s-1}, L_{s-1}, L_{s-2}$).  The motivation for doing this is to  combine  univariate  time series prediction approach (moving average) with multivariate regression approach (surrogate model) for robust estimation where we take both current and past information  into account. In Stage 1.6,  we combine the likelihood moving average  with the pseudo-likelihood to give a prediction that considers the present replica proposal and the past behavior, $L_{local}$ =  (0.5 * $L_{surrogate}$) + 0.5 * $L_{past}$. \textcolor{black}{ Note that although we use the past three values for the moving average, this number can change for different types of problems.  Once the swap interval ($\phi$) is reached, Stage 2.0 prepares the replica transition in the manager process (highlighted in blue).   Stage 3.0 executes once we reach the surrogate interval where we use data collected from Stage 1.8 for creating surrogate training set batch  $\Theta$ as shown in Equation 10. Stage 4 shows how we execute the global surrogate training in the Manager process with combined surrogate data using the neural network model in Equation 1.  We use the trained knowledge from the global surrogate model ($\Psi_{global}$)  in the respective replicas local surrogate model ($\Psi_{i}$) as shown in  Stage 1.3 and Figure \ref{fig:surrogate}. Once we reach the maximum number of samples for the given replica ($R_{max}$, Stage 5.0 signals the manager process to decrement the number of replicas alive for executing the termination criterion.  }

\textcolor{black}{Finally, we execute Stage 6  in the manager process where we combine the respective replica predictions and proposals (weights and biases) from the ensemble by concatenating the history of the samples to create the posterior distributions. This features the accepted samples and copies of the accepted samples in cases of rejected samples, as shown in Stage 1.9 of Algorithm 1.  
}

Furthermore, the framework features parallel tempering in the first stage of sampling that transforms into a local mode or exploitation in the second stage where the temperature ladder is changed such that $T_i = 1$, for all replicas, $i = 1, 2, ..., N$ as done in \cite{Chandra2019NC,chandra2019PT-Bayeslands}.  We emphasize on exploration in the first phase and emphasize on exploitation in the second phase, as shown in Stage 1.9.1 of Algorithm 1. The duration of each phase is problem dependent, which we determine from trial experiments.

We validate the quality of the surrogate model prediction using the root mean squared error (RMSE) 

$$RMSE = \sqrt{\frac{1}{N} \sum_{i=1}^{N} (L_i - \hat{L_i})^2}$$

\noindent where $L_i$ and $\hat{L_i}$ are the true likelihood and the pseudo-likelihood values, respectively. $N$ is the number of cases we employ the surrogate during sampling.

 Note that instead of swapping entire replica configuration, an alternative is to swap the respective temperatures.   This could save the amount of information exchanged during inter-process communication as the temperature is only a double-precision number, implemented by \cite{Ray2019}.


%
\begin{algorithm*} 
\smaller
 \KwData{Classification Dataset}
 \KwResult{Posterior distributions for neural network weights} 
 \tikzmk{A} {
 * Set the  number of replicas ($M$) in ensemble as $alive$;  $alive = M$ \\
 
 * Define geometric() temperature ladder , number of replica processes ($M$), surrogate interval ($\psi$), replica swap interval ($R_{swap}$),  and maximum number of samples for each replica ($R_{max}$).\\
  }\tikzmk{B}
 \boxit{blue}
  \tikzmk{A}  \While{ ($ alive \neq 0 $}{ 
 Stage 0: Prepare manager process to execute each replica in  parallel cores \\ 
 \tikzmk{B}
 \boxit{blue}
\tikzmk{A}\For{each  $i$ until $M$}{
{ $s = 0$\\
first-phase: $T_i =$ geometric() \\
  \While{ ($s < R_{max} $)}{ 
 Stage 1.0: Metropolis Transition\\  
 \For{each  $v$ until $\psi$}{
  \For{each  $k$ until $R_{swap}$}{
  1.1  Random-walk, $\theta^*_s = \theta_s + \epsilon$\\
  1.2  $L_{local}$ calculate:\\
   Draw $\kappa$ from a Uniform distribution [0,1] \\
  \uIf{$\kappa <$ $S_{prob}$ and $s > $  $\psi$ }{Estimate $L_{local}$ from local surrogate's prediction, $L_{surrogate}$ \\
  1.3 Copy global surrogate  knowledge to local surrogate, $\Psi_{i} \leftarrow \Psi_{global}$\\
  1.4 Predict $L_{surrogate}$ value with the proposed $\theta^*_i$ \\
  1.5 $L_{past}$ = mean($L_{s-1}, L_{s-1}, L_{s-2}$)\\
  1.6 Assign $L_{local}$ =  (0.5 * $L_{surrogate}$) + 0.5 * $L_{past}$ \\ 
} 
  
  \Else{ 1.7 $L_{local}$ = true-likelihood, given by 
  Likelihood function in Equation \eqref{multinomial})  \\ 
    1.8 Save $L_s$ = $L_{local}$ (Equation 10) \\}
  1.9 Calculate acceptance probability $\alpha$ and  draw $u$  from uniform distribution  \\
  
  \If{$u  \leq  \alpha $}{
  Accept replica state, $\theta_s \leftarrow \theta^*_s$
  }
  \Else{Reject and retain previous state: $\theta_s \leftarrow \theta^*_{s-1}$}
  
  1.9.1 Switch sampling style by updating replica temperature \\
  \If{second-phase is true}{
  Update temperature, $T_i = 1$\\
  }

    Increment $s$
   
  }
  
    \tikzmk{B}
 \boxit{pink}
  \tikzmk{A}
  Stage 2.0: Replica Transition: \\
  2.1  Calculate acceptance probability $\beta$   and  draw $b$ from a Uniform distribution [0,1]\\
  \If{$ b  \leq  \beta $}{
  
  2.2 Signal() manager  process  
  2.3 Exchange neighboring Replica, $\theta_i \leftrightarrow \theta_{s+1}  $
  }
   \tikzmk{B}
  \boxit{blue}
  }
  \tikzmk{A}
   Stage 3.0: Set  $\Theta_i$ which features history of proposals $\Phi$ ($\theta$) and response $ \lambda$ ( $L_{local}$ ). Use data collected from Stage 1.8

      \tikzmk{B}
  \boxit{pink}

 \tikzmk{A} Stage 4.0: Global Surrogate Training\\
 
  \For{each replica}{
  4.1 Get replica  surrogate data $\Theta_i$ from Stage 3.0\\
 
  } 
  4.2 Train global  surrogate model with combined surrogate data, $\Upsilon = [\Theta_1, \Theta_2, ... \Theta_N]$ using neural network model in Equation 1\\
  4.3 Save global surrogate model parameters, $\Psi_{global}$\\
  }
  \tikzmk{B}
 \boxit{blue}
  }

 \tikzmk{A}
   Stage 5.0: Signal() manager  process   \\
    \tikzmk{B}
 \boxit{pink}
   \tikzmk{A}
   5.1 decrement number of replica processes $alive$ \\
    \tikzmk{B}
 \boxit{blue}
   }}
   \tikzmk{A}
 Stage 6: Combine predictions and posterior from respective replicas in the ensemble,  using second-phase MCMC samples. \\
   \tikzmk{B}
 \boxit{blue}

 \caption{\small Surrogate-assisted parallel tempering for Bayesian neural networks. The highlighted regions of the algorithm shows different processing cores. The manager process is highlighted in blue while the ensemble of replica processes running in parallel is highlighted in pink.}
 \label{alg:ptfnn}
\end{algorithm*}


\subsection{Langevin gradient-based proposal distribution}

Apart from random-walk proposals, we utilize \textit{stochastic gradient Langevin dynamics} \cite{welling2011bayesian} for the proposal distribution. It features additional noise with stochastic gradients to optimize a differentiable objective function which has been very promising for neural networks \cite{BayesianChandraAC17,Chandra2019NC}.  The proposal distribution is constructed as follows.

\begin{eqnarray}
\boldsymbol\theta^p &\sim& \mathcal{N}(\bar{\boldsymbol\theta}^{[k]},\Sigma_{\theta}),\;\mbox{where}\;\label{update}\\
\bar{\boldsymbol\theta}^{[k]}&=&\boldsymbol\theta^{[k]} +r\times\nabla E_{\bf y_{\mathcal{A}_{D,T}}}[\boldsymbol\theta^{[k]} ],\label{gradient} \nonumber\\
 E_{\bf y_{\mathcal{A}_{D,T}}}[\boldsymbol\theta^{[k]}]& = & \sum_{t\in{\mathcal{A}_{D,T}}}(y_t-f({\bf x}_t)^{[k]})^2,\nonumber\\
\nabla E_{\bf y_{\mathcal{A}_{D,T}}}[\boldsymbol\theta^{[k]} ]&=& 
\left(\frac{\partial{E}}{\partial{\theta_1}},\ldots, 
\frac{\partial{E}}{\partial{\theta_{L}}}\right)\nonumber
\end{eqnarray}
$r$ is the learning rate, $\Sigma_{\theta}=\sigma^2_{\theta}I_{L}$ and $I_L$ is 
the $L \times L$ identity matrix. So that the newly proposed value of 
$\boldsymbol\theta^p$, consists of 2 parts: \begin{enumerate}
 \item  An gradient descent based weight update given by Equation  11.
 \item Add an amount of noise, from $\mathcal{N}(0,\Sigma_{\theta})$.  
 \end{enumerate}

\begin{figure}[htp!]
\includegraphics[width=90mm]{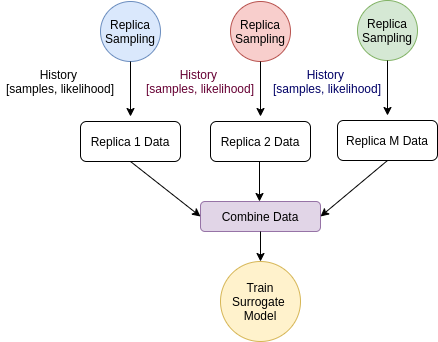}
 \caption{Data collection for training surrogate model  }
 \label{fig:surrogate_data}
\end{figure}

 \subsection{Surrogate model}

\begin{figure*}[htp!]
\includegraphics[width=\textwidth]{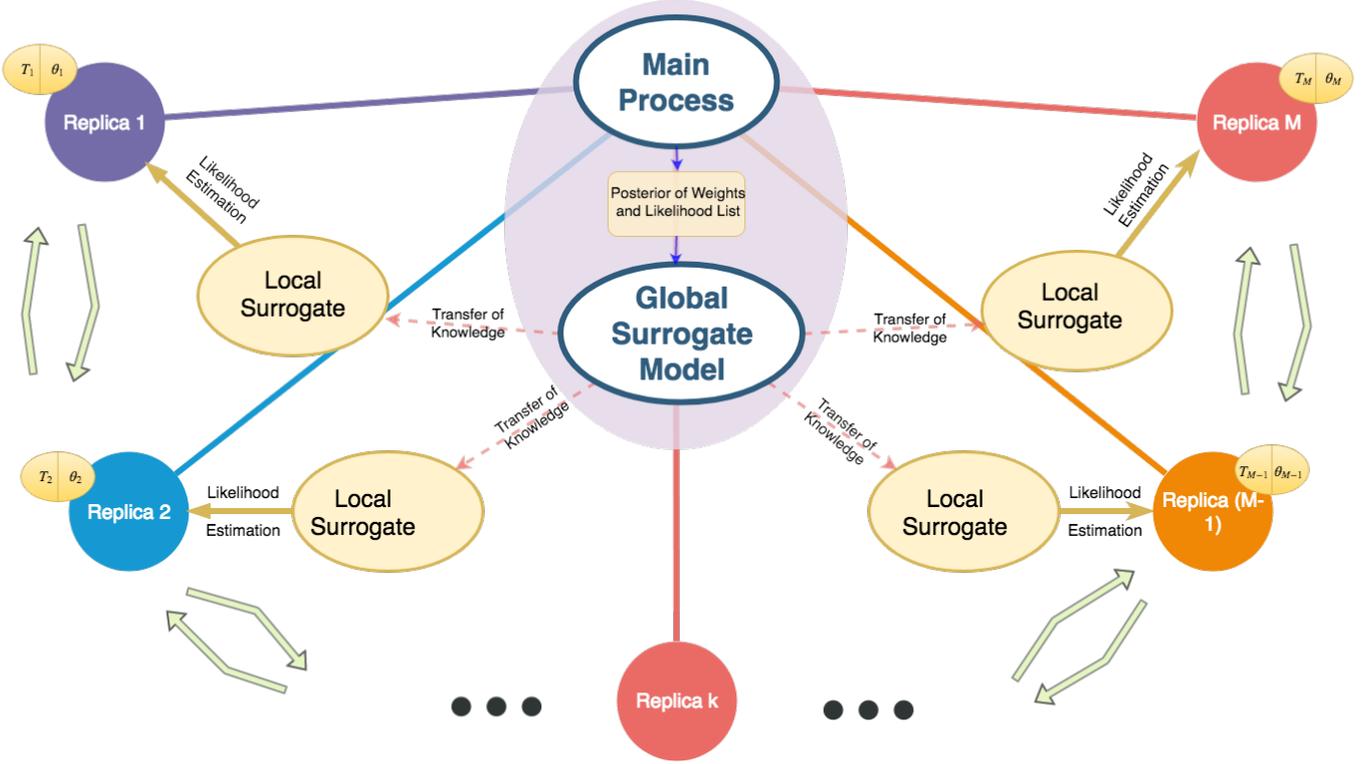}
 \caption{Surrogate-assisted multi-core parallel tempering  features surrogates to estimate the   likelihood function at times  rather than evaluating it.   }
 \label{fig:surrogate}
\end{figure*}

The choice of the surrogate model needs to consider the computational resources taken for training the model during the sampling process. We note that Gaussian process models, neural networks, and radial basis function models \cite{broomhead1988radial} have been popular choices for surrogates in the literature. 
 
 In our case, we consider the inference problem that features hundreds to thousands of parameters; hence, the model needs to be efficiently trained without taking lots of computational resources. Moreover, the flexibility of the model to have incremental training is also needed.   Therefore, we rule out Gaussian process models since they have imitations given large training datasets \cite{rasmussen2004gaussian}.  In our case,  tens of thousands of samples could make the training data.   Therefore, we use neural networks as the choice of the surrogate model. The training data and neural network model can be formulated as follows. 
 
The data given to the surrogate model is $\Phi$ and $\lambda$ as in \eqref{data}, where $\Phi$ is the input and $\lambda$ is the desired output of the model. The prediction of the model is denoted by $\hat{\lambda}$. We explain the surrogate models used in the paper as follows.
 
 In our surrogate model, we consider a single hidden layer feedforward  neural network as shown in Equation \eqref{expected_y}. The main difference is that we        use  the rectified linear unitary activation function, $g(.)$.  In this case, after forward propagation,  the errors in the estimates are evaluated using the cross-entropy cost function \cite{deBoer2005} which is formulated as:

\begin{equation}
J(\bf{W,b}) = -\frac{1}{\psi}\sum\limits_{i=0}^{\psi} (\lambda^{(i)}log(\hat{\lambda}^{(i)}) + (1-\lambda^{(i)})log(1-\hat{\lambda}^{(i)}) 
\end{equation}

\noindent The learning or optimization task then is to iteratively update the weights and biases to minimize the cross-entropy loss, $J(\bf{W,b})$.

We note that stochastic gradient descent maintains a single learning rate for all weight updates and typically the learning rate does not change during training. 
Adam (adaptive moment estimation) learning algorithm \cite{kingma2014adam} differs from classical stochastic gradient descent, as the learning rate is maintained for each network weight and separately adapted as learning unfolds. Adam computes individual adaptive learning rates for different parameters from estimates of first and second moments of the gradients. Adam features the strengths of  \textit{root mean square propagation} (RMSProp) \cite{tieleman2012lecture},  and \textit{adaptive gradient algorithm} (AdaGrad) \cite{duchi2011adaptive}.  Adam has shown better results when compared to stochastic gradient descent,  RMSprop and AdaGrad. Hence, we consider Adam as the designated algorithm for the neural network-based surrogate model.

We note that the surrogate model predictions do not provide uncertainty quantification since they are not probabilistic models.  The proposed framework is general and hence different surrogate types of models can be used in future.   Different types of surrogate models would consume different computational time and have certain strengths and weaknesses \cite{jin2001comparative}.

 \section{Experiments and Results}

 In this section, we present an experimental analysis of surrogate-assisted parallel tempering (SAPT) for Bayesian neural learning. The experiments consider a wide range of issues that test the accuracy of pseudo-likelihood by the surrogate, the quality in decision making given by the classification performance, and the amount of computational time saved. 
 
\subsection{Experimental Design}

We select six benchmark pattern classification problems from the \textit{University of California Irvine} machine learning repository \cite{Dua2019}. The problems feature different levels of computational complexity and learning difficultly;  in terms of the number of instances, the number of attributes, and the number of classes, as shown in Table \ref{tab:datas}. We use the multinomial likelihood given in Equation \ref{multinomial} the selected classification problems.  
Moreover, we show the performance using Langevin-gradient proposals which takes more computational time due to cost of computing gradients when compared to random-walk proposals; however,  it gives better prediction accuracy given the same number of samples   \cite{Chandra2019NC}. The experimental design follows the following strategy in evaluating the performance of the selected parameters from Algorithm \ref{alg:ptfnn}.

\begin{itemize}
\item Evaluate the effect of the surrogate probability ($S_{prob}$ on the computational time and classification performance. 

\item Evaluate the effect of the surrogate interval ($\psi$ on the computational time and classification performance.

\item Evaluate the effect of Langevin-gradients for proposals in parallel tempering MCMC.

\end{itemize}.

\begin{table}[h]
\centering
\small
 \caption{ Dataset  description \cite{Dua2019}}
\label{tab:datas}
\begin{tabular}{lllll}
\hline
 \hline
Dataset&Instances&Attributes&Classes&Hidden Units\\
\hline
\hline
Iris&150&4&3&12\\
Ionosphere&351&34&2&50\\ 
 Cancer&569&9&2&12\\
Bank  &11162&20&2&50\\ 
Pen-Digit& 10992 & 16& 10 & 30 \\ 
Chess& 28056 & 6& 18& 25 \\

\hline
\end{tabular}
\end{table}

We provide the parameter setting for the respective experiments as follows. A \textit{burn-in} time    $R_{burn} = 0.50   $ (50 \%) of the samples for the respective  replica. \textcolor{black}{The burn-in strategy is standard practice for MCMC sampling  which ensures that the chain enters  a high probability region,  where the states of the Markov chain are more representative of the posterior distribution.}   Although MCMC methods more commonly use burn-in time of 10-20 \% in the literature, we use  50 \%  for getting prediction performance of preferred accuracy. The maximum sampling time,  $F_{max}  = 50,000 $ for all the respective problems, and  number of  replicas, $M =10$  which run on parallel processing cores.  The  other key parameters include   replica swap interval  ($R_{swap} = 50$), surrogate interval($\psi = 50$), replica sampling time  ($R_{max} = F_{max}/M$),         surrogate probability ($S_{prob}=0.25$,  $S_{prob}=0.50$) and maximum temperature ($T_{\max}=5$).  We determined the given values for the parameters in trial experiments. We selected  the maximum temperature in trial experiments by taking into account  the performance accuracy given with a fixed number of samples.   We provide details for the pattern classification datasets with details of Bayesian neural network topology (number of hidden units) in Table \ref{tab:datas}.  
 
In the case of   random-walk proposal distribution, we  add a Gaussian noise to the  weights and biases of the network  from a \textcolor{black}{ normal distribution  with mean of 0 and standard deviation of  $0.025$}. The\textcolor{black}{ user defined constants } for  the priors (see Equation  \ref{eq:prior}) are set as ${\sigma}^{2} = 25, {\nu}_{1} = 0$ and ${\nu}_{2} = 0$.

In the respective experiments, we compare SAPT with parallel tempering featuring random-walk proposals (PTRW) and Langevin-gradients (PTLG) taken from the literature \cite{Chandra2019NC}. Surrogate-assisted parallel tempering also features Langevin-gradients (SAPT-LG) given in Equation \eqref{gradient}. We use a learning rate of 0.5 for computing the weight update via the gradients. Furthermore, we apply Langevin-gradient with a  probability $L_{prob} = 0.5 $, and random-walk proposal, otherwise. Note that the respective methods feature parallel tempering for the first 50 per cent of the samples that make the burn-in period. Afterwards, we use canonical MCMC where temperature $T=1$ using parallel computing environment featuring replica-exchange via interprocess communication.

\subsection{Implementation}

We employ multi-core parallel tempering for neural networks \footnote{Multi-core parallel tempering: https://github.com/sydney-machine-learning/parallel-tempering-neural-net} to implement surrogate assisted parallel tempering. We use one hidden layer in the Bayesian neural network for classification problems.

We use the \textit{Keras} machine learning  library  for implementing the surrogate  \footnote{Keras: https://keras.io/} with Adam   learning algorithm \cite{kingma2014adam}. The surrogate neural network model architecture consists of $[i, h_1, h_2, o ]$; where  $i$ refers to the number of inputs that consists of the total number of weights and biases used in the Bayesian neural network for the given problem. $h_1$ and $h_2$ refers to the first and second hidden layers, and $o$ represents the output that predicts the likelihood.  In our experiments, we use hidden units $h_1=64, h_2=16$,  for the Iris and Cancer problems. In the Ionosphere and Bank problems,  we use hidden units $h_1=120, h_2=40$, and for  Pen-Digit and Chess problems, we use hidden units $h_1=200, h_2=50$.  All problems used one output unit for the surrogate model, $o=1$.

\subsection{Results}
 
We first present the results in terms of \textcolor{black}{classification accuracy of posterior samples} for SAPT-RW with random-walk proposal distribution  (Table \ref{tab:sapt-rw}) where we evaluate different combinations of selected values of surrogate interval and $\psi$    surrogate probability  $S_{prob}$.   Looking at the elapsed time, we find that SAPT-RW is more costly for the Iris and Cancer problems when compared to PT-RW. These are smaller problems when compared to rest given the size of the dataset shown in Table \ref{tab:datas}. The Ionosphere problem saved computation time for both instances of SAPT-RW.  A larger dataset with a Bayesian neural network model implies that there is more chance for the surrogate to save time which is visible in the   Pen-Digit and Chess problems. The Bank problem does not save much time but retains the accuracy in classification performance. Furthermore, we find that instances of SAPT improve the classification accuracy of Iris, Ionosphere, Pen-Digit and Chess problems. In Cancer and Bank problems, the performance is similar. 
 
 We provide the results for Langevin-gradient proposals (SAPT-LG) in Table \ref{tab:result_lg}. In general, the classification performance improves when compared to random-walk proposal distribution (SAPT-RW) in Table \ref{tab:sapt-rw}. \textcolor{black}{By incorporating the gradient into the proposal, the results show faster convergence with better accuracy given the same number of samples. In a comparison of both forms of parallel tempering (PT-RW and PT-LG) with the surrogate-assisted framework (SAPT-RW and SAPT-LG), we observe that the elapsed time has not improved for Iris, Ionosphere and Cancer problems; however, it has improved in the rest of the problems. }
 
 The accuracy of the surrogate in predicting the likelihood is shown in Table \ref{tab:surr_acc} for the smaller problems that feature Ionosphere, Cancer and Iris. We notice that the RMSE for surrogate prediction is lower for the Iris problem when compared to the others; however, as shown in Figure \ref{fig:iris-cancer}, Figure \ref{fig:ionos-bank}  and \ref{fig:pendigit-chess}, this is relative to the range of log-likelihood. We observe that the log-likelihood prediction by the surrogate model is much better for the smaller problems (Iris and Cancer) when compared to the larger problems (Pen-Digit and Chess).

 \textcolor{black}{Figure \label{fig:pos} shows the comparison of the posterior distribution of two selected weights from input to the hidden layer of the Bayesian neural network for the Cancer problem using SAPT-RW and PT-RW. We notice that although the chains converged to different sub-optimal models in the multimodal distributions, SAPT-RW does not depreciate in terms of exploration.}

\begin{figure*}[htp!]
\centering
    \begin{tabular}{cc} 
      \subfigure[Weight 1  ]{\includegraphics[width=70mm]{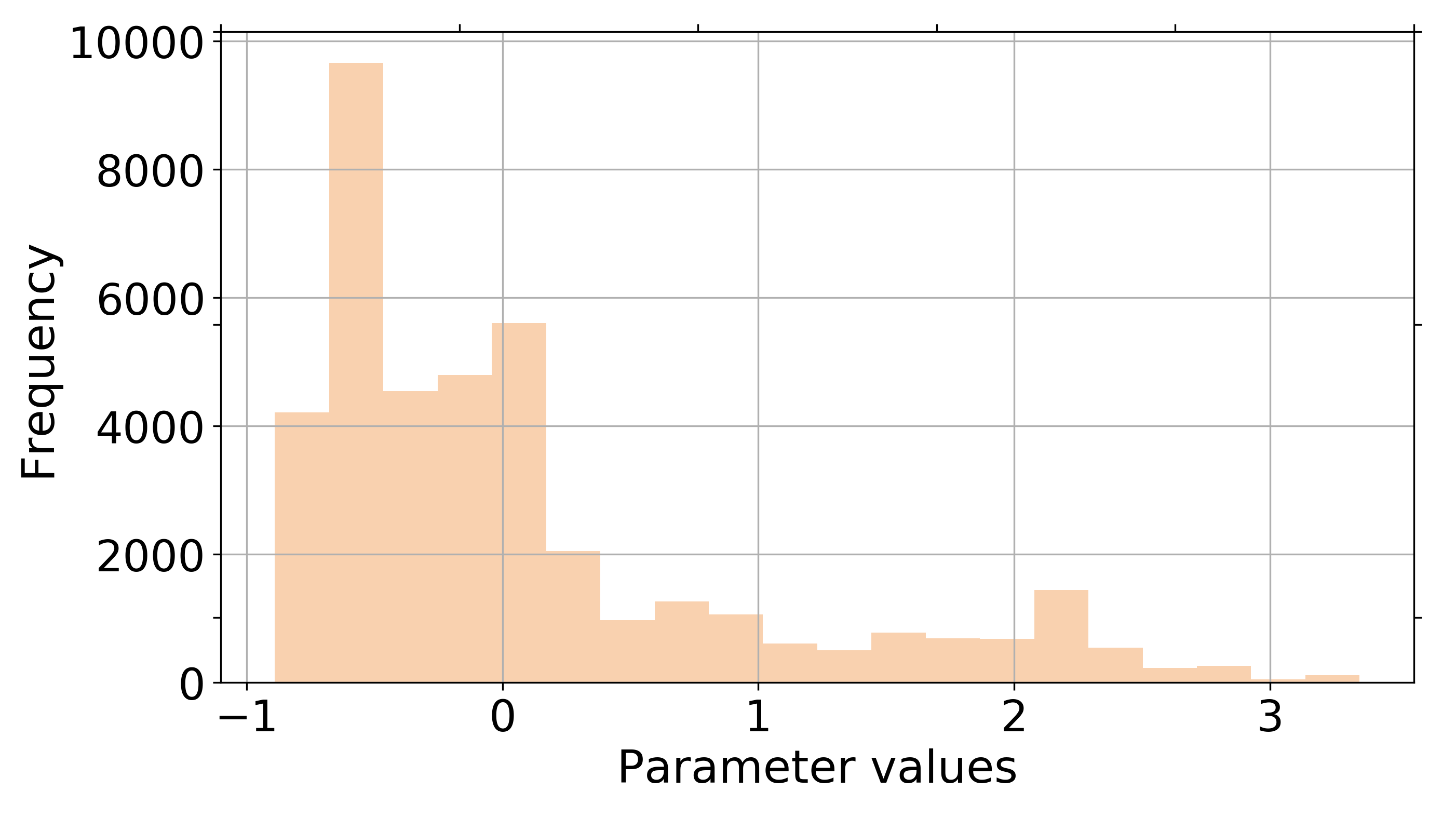}}
      \subfigure[Weight 1 ]{\includegraphics[width=70mm]{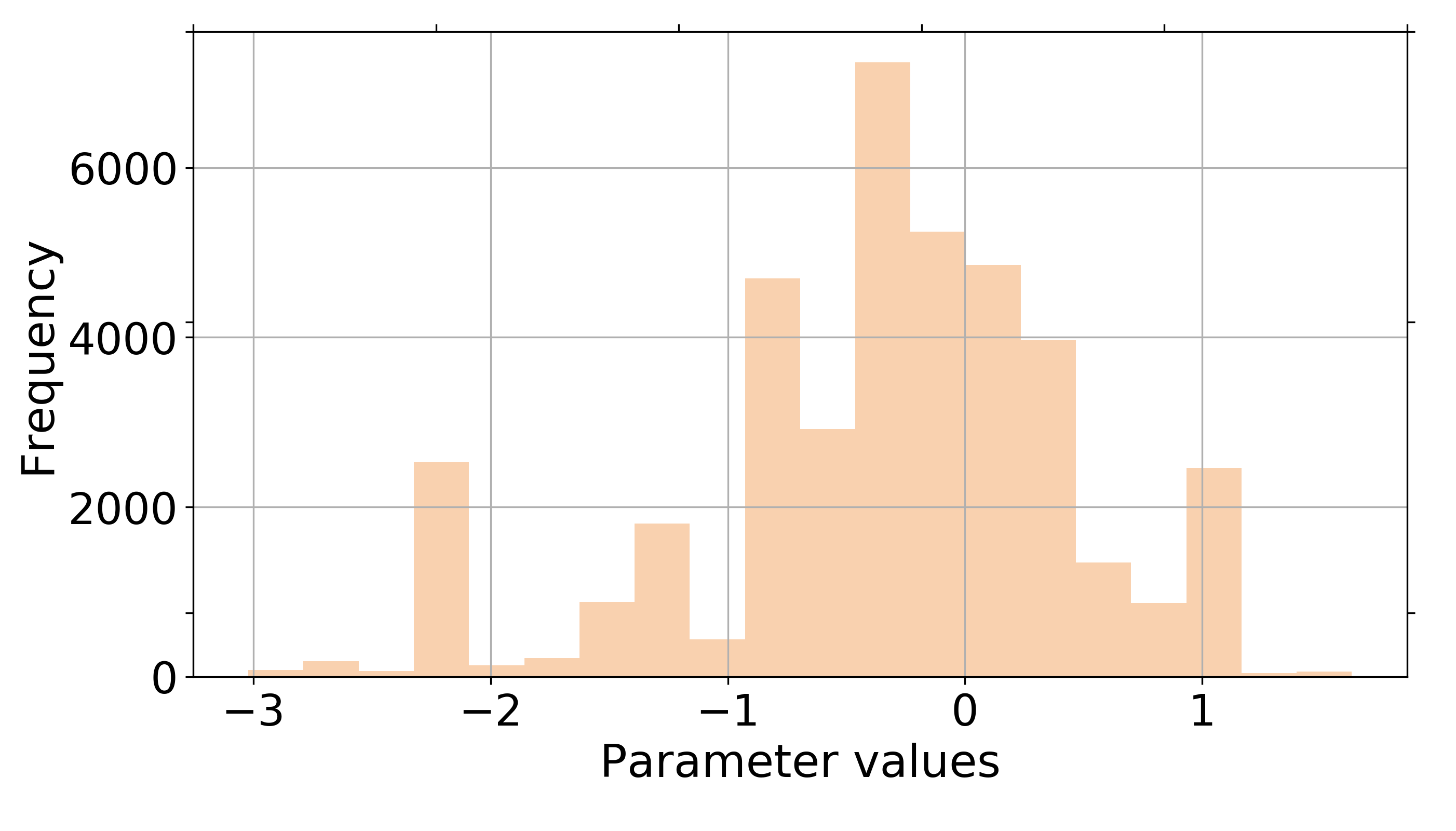}}\\
      \subfigure[Weight 2 ]{\includegraphics[width=70mm]{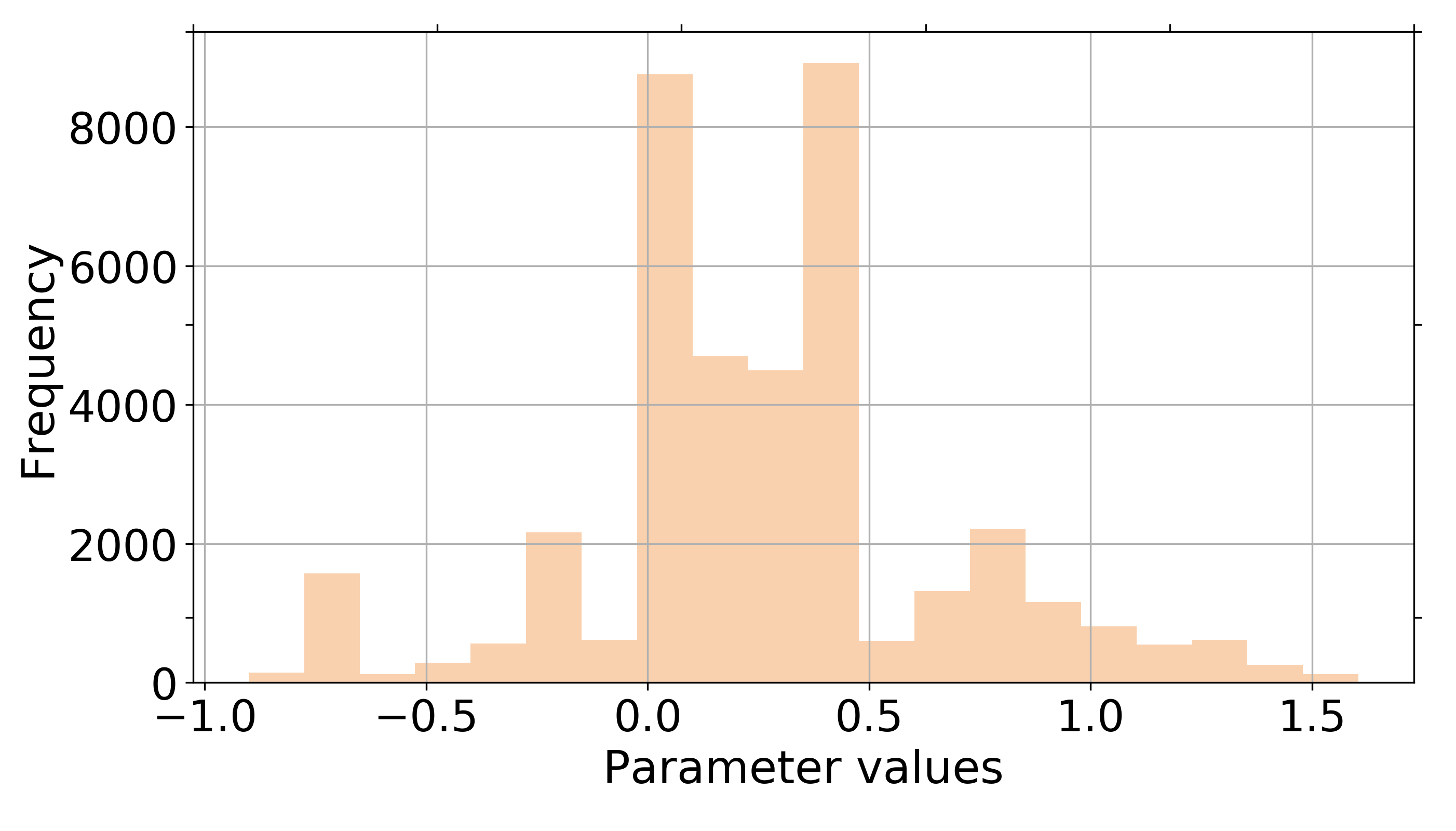} }      
      \subfigure[Weight 2 ]{\includegraphics[width=70mm]{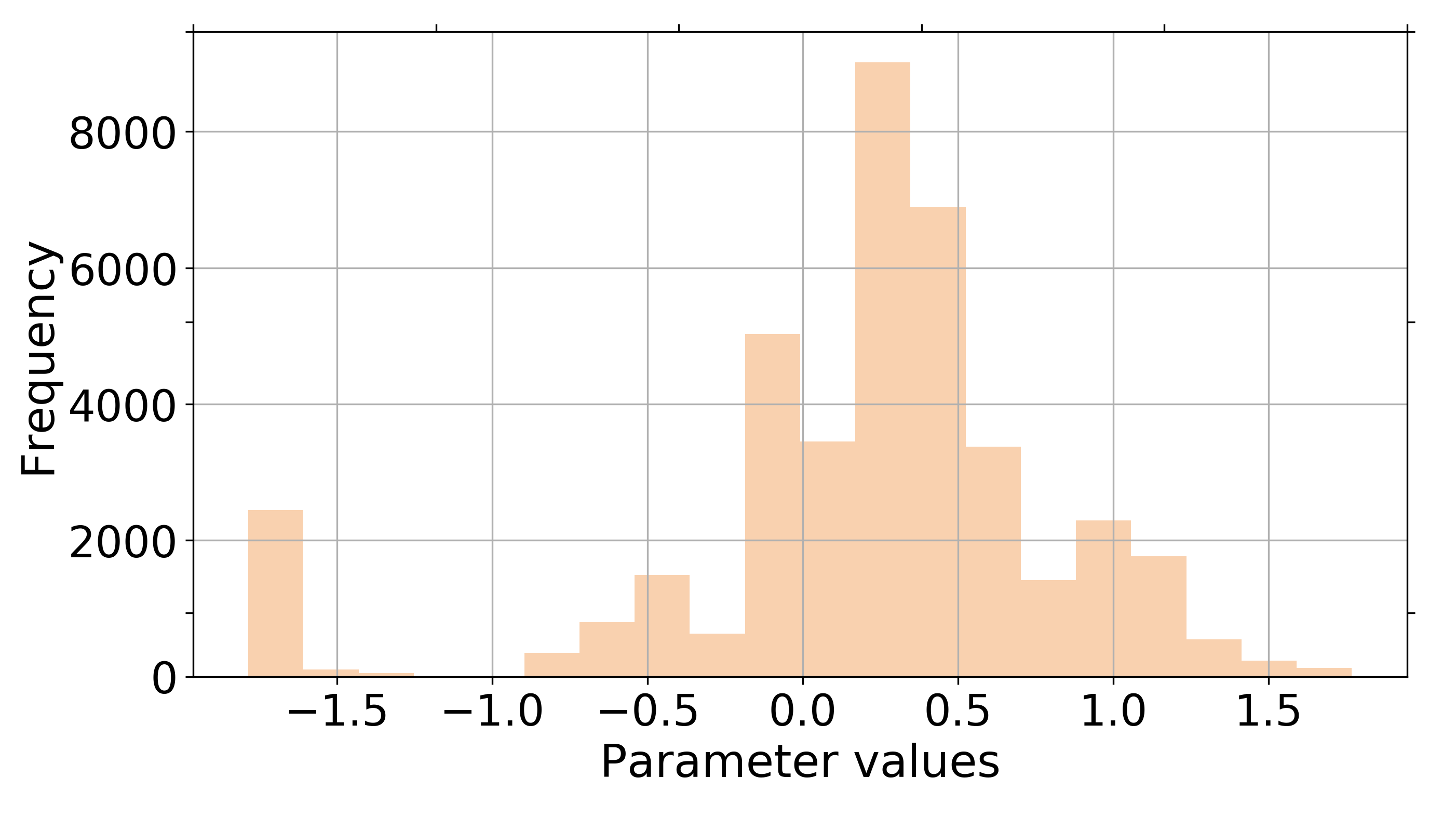} }      
       
    \end{tabular}
 
\caption{Posterior distribution and trace-plot  comparing surrogate-assisted parallel tempering (Panel (a) and (c))  with parallel tempering (Panel (b) and (d))  MCMC  for selected weights (1 and 2) for the Cancer problem.  }
\label{fig:pos}
\end{figure*}

\begin{figure}[htp!]
\centering
\begin{subfigure}
\centering
\includegraphics[width=90mm]{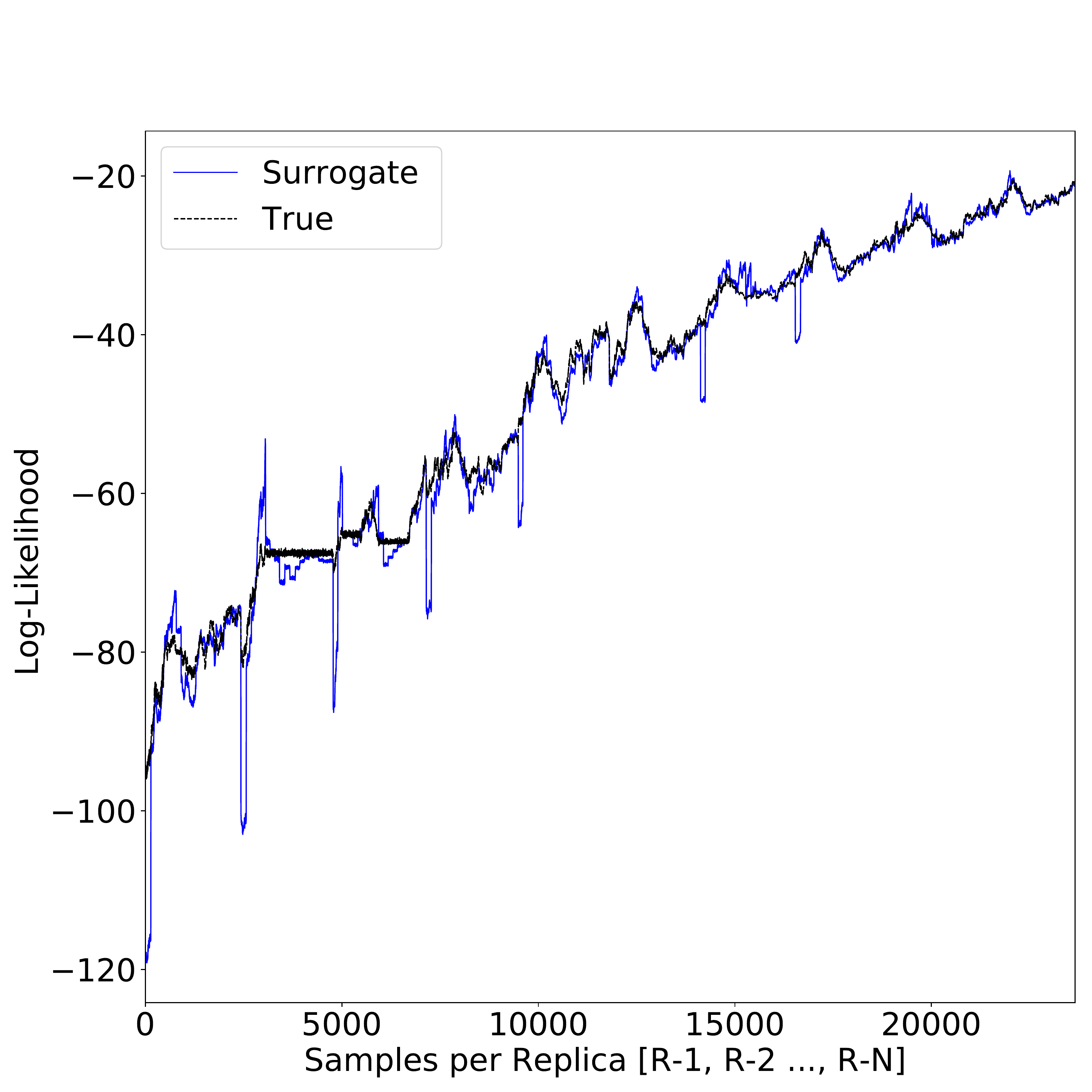}
\end{subfigure}
\begin{subfigure}
\centering
\includegraphics[width=90mm]{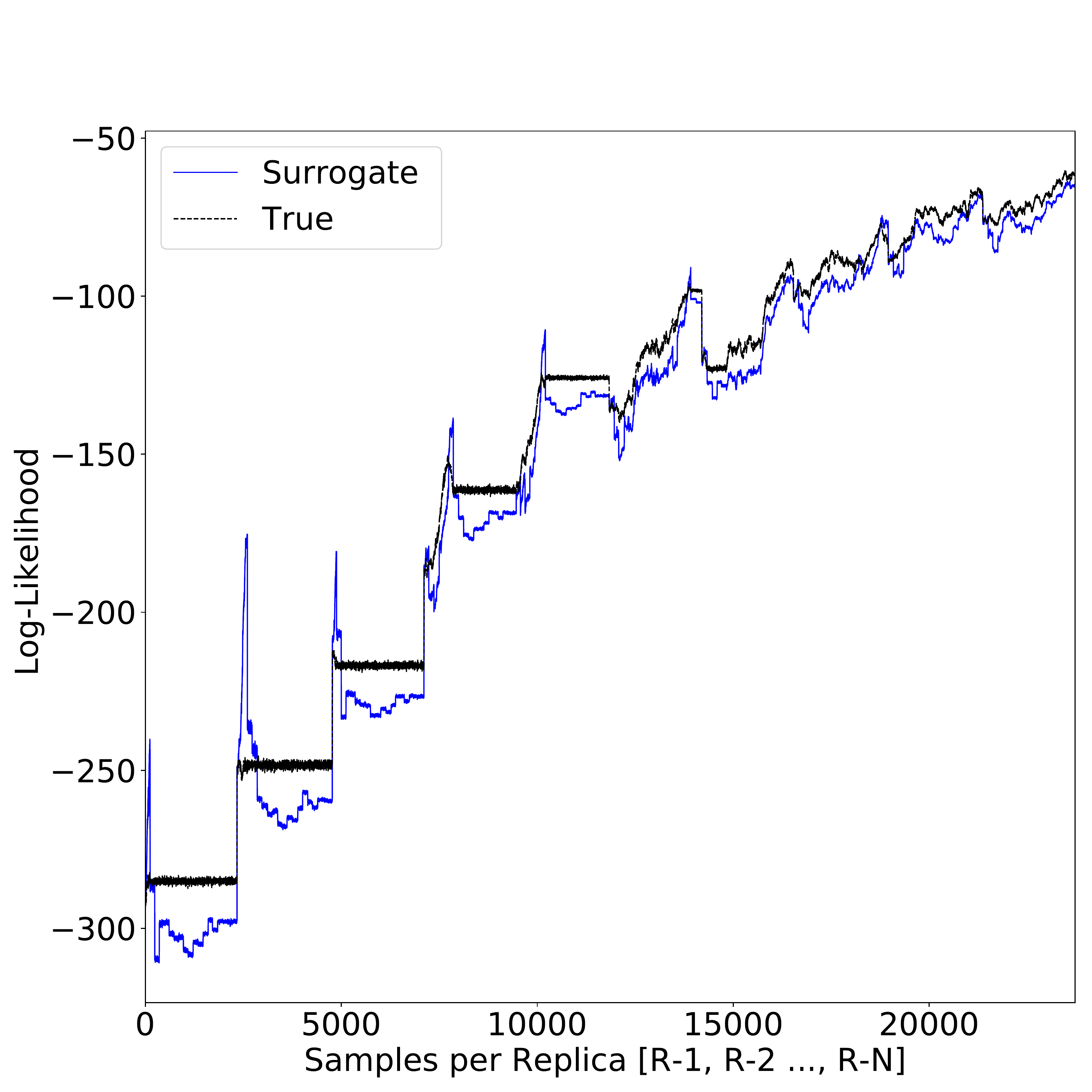}
\end{subfigure}
\caption{ The Iris (top) and Cancer (bottom) surrogate accuracy. The dashed line denotes the   real likelihood function while the blue line gives the surrogate likelihood estimation.}
\label{fig:iris-cancer}
\end{figure}

\begin{figure}[htp!]
\centering
\begin{subfigure}
\centering
\includegraphics[width=90mm]{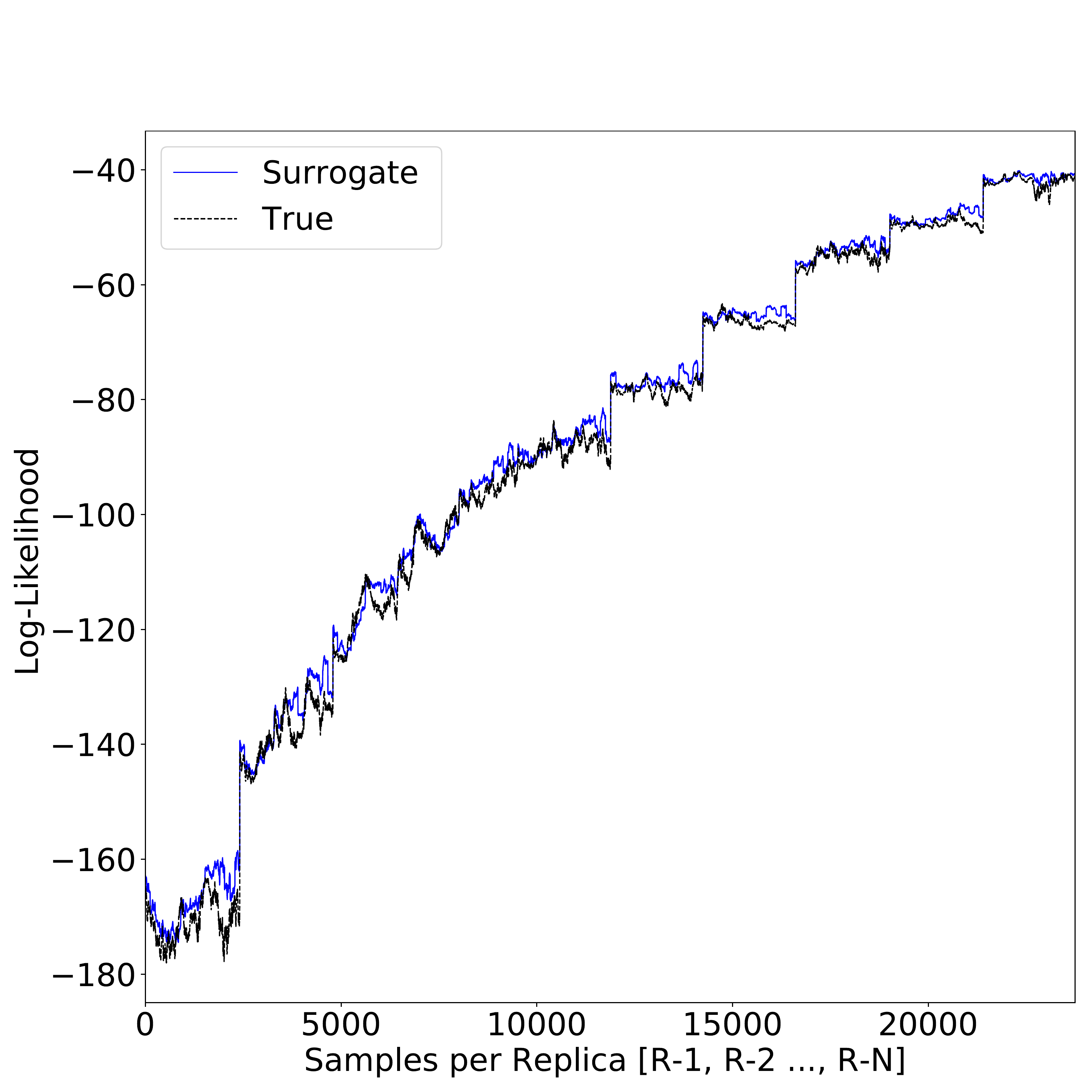}
\end{subfigure}
\begin{subfigure}
\centering
\includegraphics[width=90mm]{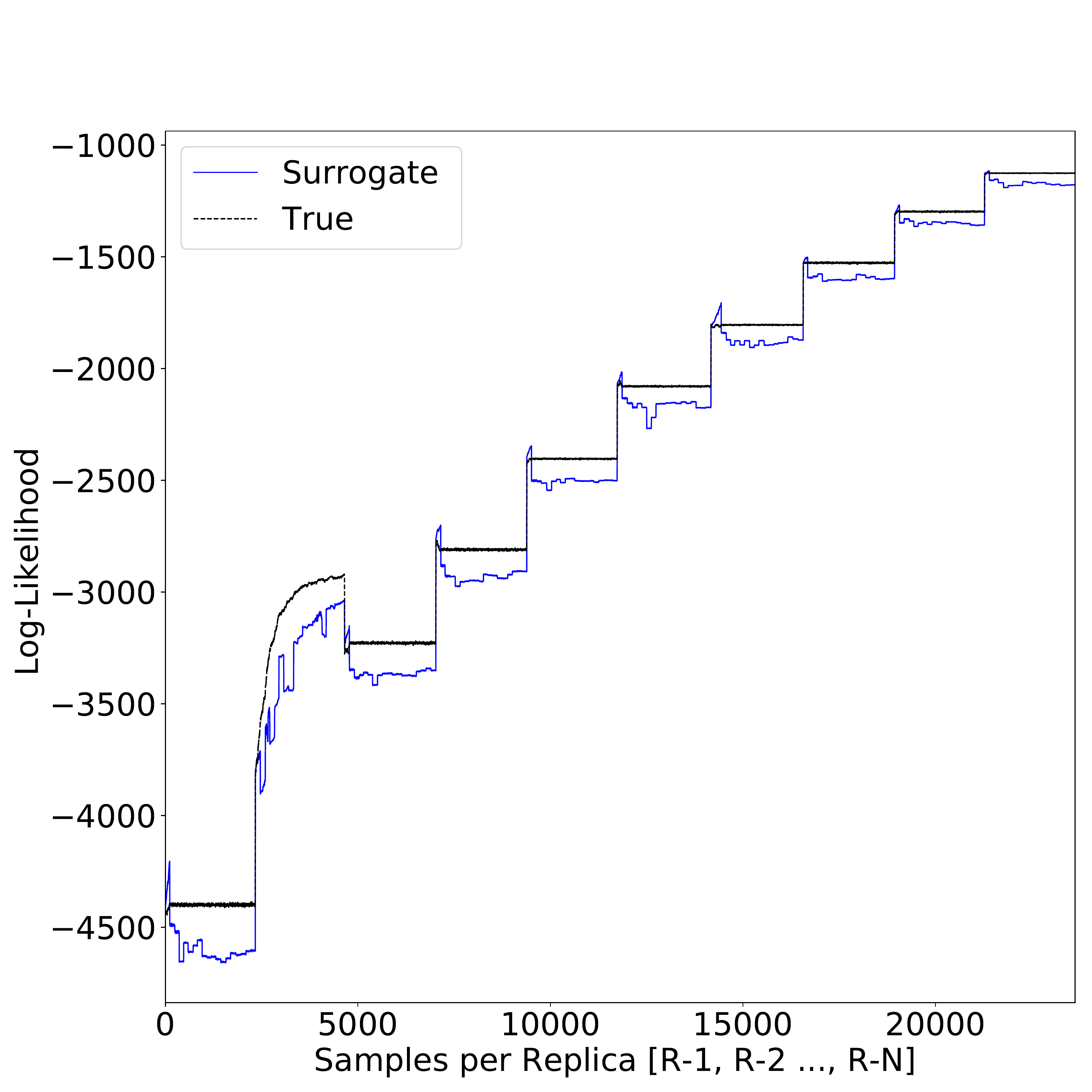}
\end{subfigure}
\caption{The Ionosphere (top) and Bank   (bottom) surrogate accuracy. The dashed line denotes the   real likelihood function while the blue line gives the surrogate likelihood estimation.}
\label{fig:ionos-bank}
\end{figure}

\begin{figure}[htp!]
\centering
\begin{subfigure}
\centering
\includegraphics[width=90mm]{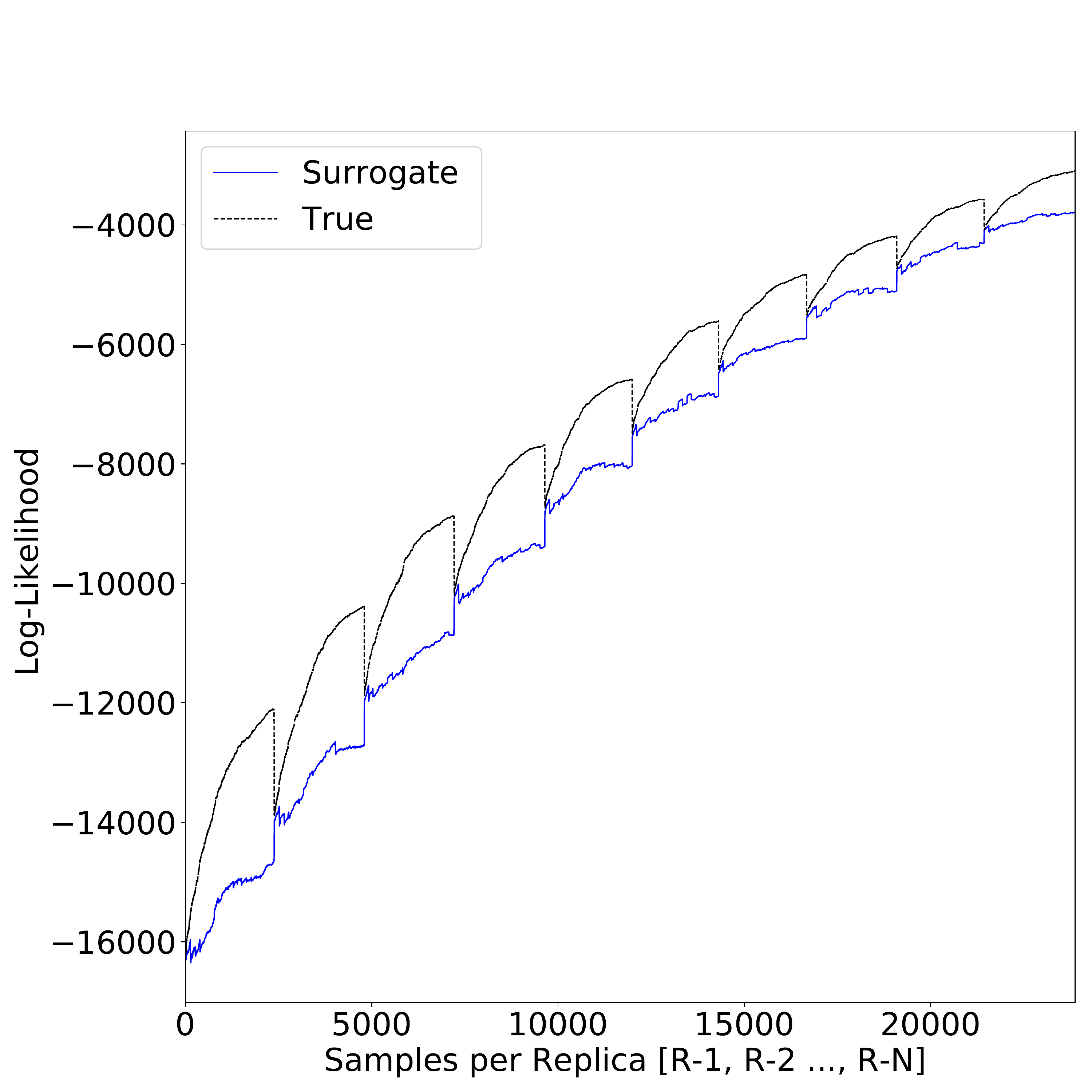}
\end{subfigure}
\begin{subfigure}
\centering
\includegraphics[width=90mm]{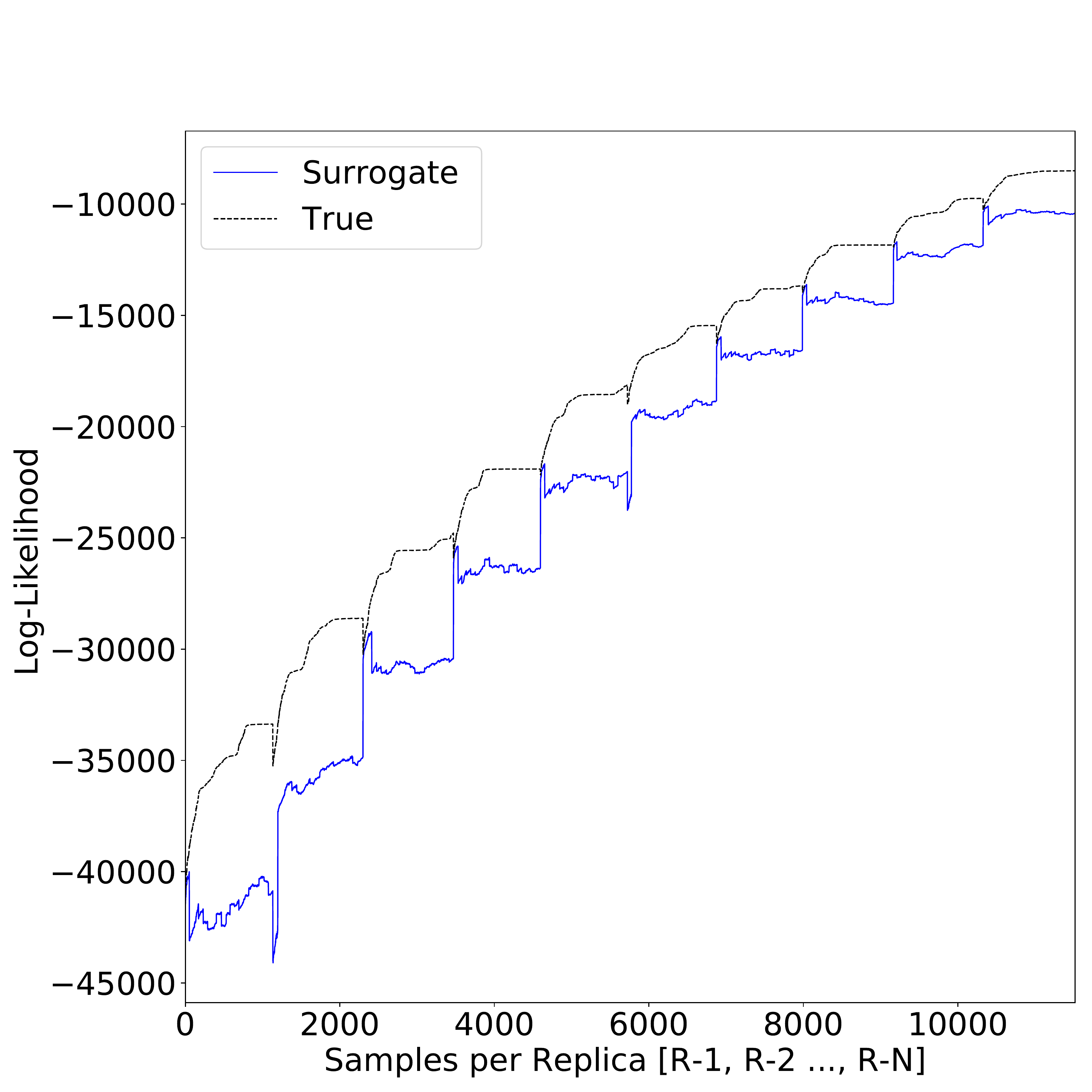}
\end{subfigure}
\caption{The Pen-Digit (top) and Chess (bottom) surrogate accuracy. The dashed line denotes the   real likelihood function while the blue line gives the surrogate likelihood estimation.}
\label{fig:pendigit-chess}
\end{figure}

\begin{table*}[h]
\centering
\small
 \caption{Classification Results (Random-walk proposal distribution)}
\label{tab:sapt-rw}
\begin{tabular}{llllll}
\hline
 \hline
Dataset&Method&Train Accuracy&Test Accuracy& Elapsed Time &\\

 &    SAPT($S_{prob}$)  
&[mean, std, best] & [mean, std, best]& (minutes) & \\
\hline
\hline
 Iris&PT-RW& 51.39 15.02 91.43 & 50.18 41.78 100.00 &1.26\\
 
&SAPT-RW (0.25)& 69.31 2.05 80.00 &65.24 4.04 78.38 & 1.93    \\ 
 &SAPT-RW (0.50) & 44.11 11.93 78.89& 50.01 5.65 70.30 & 1.81  \\
\hline

Ionosphere&PT-RW& 68.92 16.53 91.84& 51.29 30.73 91.74 & 3.50\\
&SAPT-RW (0.25) & 76.60 4.42 86.94 & 61.23 8.05 86.24  & 2.93  \\
&SAPT-RW (0.50) & 70.34 6.15 83.27 & 73.42 14.06 95.41  & 2.43    \\
\hline
Cancer&PT-RW& 83.78 20.79 97.14 &83.55 27.85 99.52&2.78\\
&SAPT-RW(0.25)&89.75 6.91 96.32 & 92.80 4.57 99.05  & 3.41  \\
&SAPT-RW(0.50)&91.17 6.29 96.52 & 97.64 3.17 99.52  & 2.84  \\
\hline
Bank  &PT-RW&78.39 1.34 80.11& 77.49 0.90 79.45 &  27.71\\  
&SAPT-RW(0.25)&  78.44 0.67 79.69 &77.79 0.63 79.60 & 28.67   \\

&SAPT(0.50)& 77.82 1.05 79.69 &77.16 0.91 78.80 & 27.38  \\
\hline
Pen Digit&PT-RW& 76.67 17.44 95.24& 71.93 16.59 90.62   &57.13 \\
 &SAPT-RW(0.25) &  88.74 1.94 92.87 & 83.60 2.14 88.74 &  49.25   \\
&SAPT-RW(0.50) & 80.85 1.28 82.87 & 77.66 1.08 80.02  & 36.05  \\
\hline
Chess&PT-RW  &   89.48 17.46 100.00 &90.06 15.93 100.00 &252.56 \\
 &SAPT-RW(0.25) &   97.17 8.35 100.00 & 97.66 6.83 100.00 &197.61  \\ 
&SAPT-RW(0.50) &     90.87 13.35 100.00 & 90.71 13.31 100.00  & 143.75   \\
\hline
\hline
\end{tabular}
\end{table*}

\begin{table*}[h]
\centering
\small
 \caption{Classification Results (Langevin-gradient proposal distribution)}
\label{tab:result_lg}
\begin{tabular}{llllll}
\hline

 \hline
Dataset&Method&Train Accuracy&Test Accuracy& Elapsed Time &\\

 & &[mean, std, best] & [mean, std, best]& (minutes) & \\
\hline 
\hline

Iris&PT-LG&  97.32 0.92 99.05 &96.76 0.96 99.10 & 2.09 \\
&SAPT-LG (0.25) &98.91 0.16 100.00& 99.93 0.39 100.00 &2.85\\ 
  &SAPT-LG (0.50) & 99.09 0.43 100.00 &98.63 1.57 100.00 &  2.49\\ 
\hline
Ionosphere&PT-LG& 98.55 0.55 99.59 &92.19 2.92 98.17 &5.07  \\
  &SAPT-LG (0.25 ) &100.00 0.02 100.00& 90.82 2.43 96.33 & 6.17\\
 &SAPT-LG (0.50 ) & 99.51 0.60 100.00& 91.24 1.82 97.25 & 4.76 \\
\hline
Cancer&PT-LG& 97.00 0.29 97.75 &98.77 0.32 99.52&5.09  \\
 &SAPT-LG(0.25)&99.36 0.11 99.39& 98.00 0.76 99.52 & 8.18 \\
 &SAPT-LG(0.50)&99.37 0.12 99.59 &98.61 0.65 100.00 & 6.64  \\
\hline
Bank &PT-LG& 80.75 1.45 85.41 &79.96   0.81 82.61 & 86.94\\  
 &SAPT-LG(0.25)& 79.86 0.15 80.30 & 80.53 0.28 79.22 &75.96 \\  
 &SAPT-LG(0.50)& 80.86 0.15 80.30 & 81.53 0.28 79.25 &65.11  \\  
\hline
Pen Digit&PT-LG&  84.98 7.42 96.02 & 81.24 6.82 91.25 & 86.62\\  
 &SAPT-LG(0.25)& 82.12 7.42 94.02 & 82.24 6.82 93.25 & 66.62  \\   
 &SAPT-LG(0.50)& 83.98 7.42 95.02 & 83.14 6.82 92.25 & 56.62 \\  
\hline
Chess&PT-LG  &  100.00 0.00 100.00 & 100.00 0.00 100.00  & 323.10\\
 &SAPT-LG(0.25)& 100.00 0.00 100.00 & 100.00 0.00 100.00  & 223.70 \\  
 &SAPT-LG(0.50)& 100.00 0.00 100.00 & 100.00 0.00 100.00  & 173.10  \\  
\hline
\hline
\end{tabular}
\end{table*}

\begin{table*}[h]
\centering
\small
 \caption{Surrogate Accuracy }
\label{tab:surr_acc}
\begin{tabular}{llllll}
\hline

 \hline
Dataset&Method  &Surrogate Prediction &Surrogate Training  \\

 & &    RMSE & RMSE [mean, std] \\
\hline
\hline
\hline

Iris &SAPT-RW & 3.55 & 3.84e-05 8.40e-05   \\ 

Ionosphere& SAPT-RW  &  2.63 & 7.20e-05 1.62e-04 \\

Cancer& SAPT-RW  &     11.08 &6.12e-05 1.58e-04 \\ 
 
Bank&SAPT-RW    & 131.26 &3.09e-05   1.53e-04 \\ 
  
  Pen-Digit & SAPT-RW  & 1246.60 & 3.93e-06 1.47e-05\\ 
   Chess& SAPT-RW  &  4026.44 &1.34e-06 3.58e-06 \\  
\hline
\hline
\end{tabular}
\end{table*}

\textcolor{black}{Table 
\ref{tab:num_replica} provides an evaluation of the number of replicas (cores) on the computational time (minutes) using SAPT-RW for selected problems. We observe that in general, the computational time reduces as the number of replica increases by taking advantage of parallel processing. }

\begin{table*}[h]
\centering
\small
 \caption{Effect of number of replica on the computational time (minutes) }
\label{tab:num_replica}
\begin{tabular}{lllllll}
\hline

 \hline
Num. Replica & Cancer &     Bank &    Pen-digit &     Chess\\

\hline
\hline

4& 45.66    & 79.25 &     102.28 &    253.98\\ 
 
6& 21.16    & 43.48    & 61.38    & 188.79 \\ 
  
8& 13.25    &32.57    &51.92    &185.52 \\ 
10& 9.17    &29.75 &    44.7 & 130.57 \\  
\hline
\hline
\end{tabular}
\end{table*}

 \section{Discussion}
 
 The results, in general, have shown that surrogate-assisted parallel tempering can be beneficial for larger datasets and models, demonstrated by Bayesian neural network architecture for Pen-Digit and Chess classification problems. This implies that the method would be very useful for large scale models where computational time can be lowered while maintaining performance in decision making such as classification accuracy. We observed that in general, the Langevin-gradients improves the accuracy of the results.    Although we used Bayesian neural networks to demonstrate the challenge in using computationally expensive models with large datasets, surrogate-assisted parallel tempering can be used for a wide range of models across different domains.  We note that  Langevin-gradients are limited to models where gradient information is available. In large and computationally expensive geoscientific models such as modelling landscape  \cite{chandra2019PT-Bayeslands} and modelling reef evolution \cite{JPall_BayesReef2018}, it is difficult to obtain gradients from the models; hence, random-walk and other meta-heuristics are more applicable.  
 
The proposed method employs transfer learning for the surrogate model, where we transfer the knowledge and refine the surrogate model in the forthcoming surrogate intervals with new data.  Since each replica of parallel tempering is executed on a separate processing core, inter-process communication is used for exchanging information between the different replicas. Inter-process communication is also used for collecting the history of information in terms of proposals and associated likelihood for creating training datasets for the surrogate model. The proposed method could be seen as a case for online learning that considers a sequence of predictions from previous tasks and currently available information \cite{shalev2012online}. This is because the surrogate is trained at every surrogate interval and the surrogate gives an estimation of the likelihood until the next interval is reached for further retraining based on accumulated data of proposal and true-likelihood for the previous interval.

We observe that there is systematic under-estimation of the true likelihood for some of the cases (Figures 4-6).  One reason could be that the variance is overestimated and the other is due to the global surrogate model, where training is done by one model in the \textit{manager process} and the knowledge is used in all the replicas. The surrogate accuracy depends on the nature of the problem, and the neural network model, in terms of the number of parameters in the model and the size of the dataset. We find that smaller Bayesian neural network models had good accuracy in surrogate estimation when compared to others. In future work, we can consider training surrogate model in the  local replicas which could further improve the estimation.  The global surrogate model has the advantage of combining information across the different replicas, it faces challenges of dealing with large surrogate training dataset which accumulates over sampling time. We also need to lower the time taken for the exchange of knowledge needed for decision making by the local surrogate model and the master replica.  We found that bigger problems  (such as Pen-Digit and Chess)  give further challenges for surrogate estimation. In such cases, it would be worthwhile to take a time series approach, where the history of the likelihood is treated as time series. Hence, a local surrogate could learn from the past tend of the likelihood, rather than the history of past proposals. This could help in addressing computational challenges given a large number of model parameters to be considered for surrogate training.

  Although we ruled out Gaussian process models as the choice of the surrogate model due to computational \textcolor{black}{complexity}  in training large surrogate data, we need to consider that Gaussian process models naturally account for uncertainty quantification in decision making.   Recent techniques to address the issue of training Gaussian process models for large datasets could be a way ahead in future studies \cite{moore2016fast}.  Moreover, we note that there has not been much work done in the literature that employs surrogate assisted machine learning. Most of the literature considered surrogate assisted optimization, whereas we considered  inference for machine learning problems. The results open the road to use surrogate models for machine learning. Surrogates would be helpful in case of big data problems and cases where there are inconsistencies or noisy data.  Furthermore, other optimization methods could be used in conjunction with surrogates for big data problems rather than parallel tempering.  
  
  Finally, we use the surrogate model with a certain probability in all replicas, hence the resulting distribution is only an approximation of the true posterior distribution. An interesting question would then be whether there is a way to use surrogates while not introducing an approximation error. This is trivially possible for sequential approaches; however, in parallel cases, this could be an issue since multiple replicas are used which all contribute to different levels of uncertainties. This would need to be addressed in future work in uncertainty quantification for the surrogate likelihood prediction, perhaps with the use of Gaussian process surrogate models or by introducing an error term for each surrogate model and integrating it out via MCMC sampling.  Potentially, multi-fidelity modelling where the synergy between low-fidelity and how fidelity data and models could be employed for the surrogate model \cite{peherstorfer2018survey}.
   
\section{Conclusions and Future Work}

We presented surrogate-assisted parallel tempering for implementing Bayesian inference for computationally expensive problems that harness the advantage of parallel processing. We used a Bayesian neural network model to demonstrate the effectiveness of the framework to depict computationally expensive problems.  The results from the experiments reveal that the method gives a promising performance where computational time is reduced for larger problems. 

The surrogate-based framework is flexible and can incorporate different surrogate models and be applied to problems across various domains that feature computationally expensive models which require parameter estimation and uncertainty quantification. In future work, we envision strategies to further improve the surrogate estimation for large models and parameters. A way ahead is to utilize local surrogate via time series prediction could help in alleviating the challenges. Furthermore, the framework could be applied to problems in different domains such as computationally expensive geoscientific models used for landscape evolution.

 \section*{Software and Data }
  We provide an open-source implementation of the proposed algorithm in Python along with data and sample results  \footnote{Surrogate-assisted multi-core parallel tempering: https://github.com/Sydney-machine-learning/surrogate-assisted-parallel-tempering}. 
  
  \section*{Credit Author Statement}
   Rohitash Chandra: conceptualization, methodology, experimentation, writing and supervision; Konark Jain: initial software development and experimentation; Arpit Kapoor: software development and experimentation; Ashray Aman: software development, experimentation,   and visualisation.

\section*{Acknowledgement}

The authors would like to thank Prof. Dietmar Muller and Danial Azam for discussions and support during the course of this research project. We sincerely thank the editors and anonymous reviewers for their valuable comments.

\bibliographystyle{elsarticle-num}

\bibliography{aicrg,2018,Chandra-Rohitash,Bays,sample,surrogate} 

 
\end{document}